\documentclass[conference]{IEEEtran}
\IEEEoverridecommandlockouts
\usepackage{cite}
\usepackage{amsmath,amssymb,amsfonts}
\usepackage{algorithmic}
\usepackage{graphicx}
\usepackage{textcomp}
\usepackage{xcolor}
\usepackage{natbib}
\usepackage{booktabs}
\usepackage{tgheros}
\usepackage[symbol]{footmisc}

\setcitestyle{numbers}
\setcitestyle{square}
\setcitestyle{citesep={,}}
\def\BibTeX{{\rm B\kern-.05em{\sc i\kern-.025em b}\kern-.08em
    T\kern-.1667em\lower.7ex\hbox{E}\kern-.125emX}}

\usepackage{xspace}
\usepackage{multirow}
\usepackage{rotating}
\usepackage{amsmath}
\usepackage{soul}
\usepackage{array}
\usepackage{tablefootnote}
\usepackage[symbol]{footmisc}

\usepackage{natbib}
\usepackage[backgroundcolor=white,textsize=tiny]{todonotes}

\newcommand{\linebreakand}{%
  \end{@IEEEauthorhalign}
  \hfill\mbox{}\par
  \mbox{}\hfill\begin{@IEEEauthorhalign}
}

\newcommand{\customfootnotetext}[2]{{% Group to localize change to footnote
  \renewcommand{\thefootnote}{#1}% Update footnote counter representation
  \footnotetext[0]{#2}}}% Print footnote text

\renewcommand{\vec}[1]{\mathbf{#1}} %% e.g., \vec{x}
\newcommand{\name}
{{\fontfamily{cmtt}\selectfont HungerGist}}

\renewcommand*{\thefootnote}{\fnsymbol{footnote}}

\newcommand{\fews}{\textit{Food Crisis Index}\xspace}

\begin{document}

\title{HungerGist: An Interpretable Predictive Model for Food Insecurity}

\author{
    \IEEEauthorblockN{
        Yongsu Ahn\textsuperscript{$\ddagger$}\IEEEauthorrefmark{1}, 
        Muheng Yan\textsuperscript{$\ddagger$}\IEEEauthorrefmark{1}, 
        Yu-Ru Lin\textsuperscript{$\S$}\IEEEauthorrefmark{1}, 
        Zian Wang\IEEEauthorrefmark{2}}
    \IEEEauthorblockA{\IEEEauthorrefmark{1}School of Computing and Information, University of Pittsburgh
    \\\{yongsu.ahn, muheng.yan, yurulin\}@pitt.edu}
    \IEEEauthorblockA{\IEEEauthorrefmark{2}Department of Computer Science, Stony Brook University
    \\\{ziawang\}@stonybrook.edu}
}

% \author{\IEEEauthorblockN{Yongsu Ahn\textsuperscript{\textsection}}
% % \author{\IEEEauthorblockN{Yongsu Ahn$^*$}
% \IEEEauthorblockA{\textit{School of Computing and Information} \\
% \textit{University of Pittsburgh}\\
% Pittsburgh, United States \\
% yongsu.ahn@pitt.edu}
% \and
% \IEEEauthorblockN{Muheng Yen\textsuperscript{\textsection}}
% % \IEEEauthorblockN{Muheng Yen$^*$}
% \IEEEauthorblockA{\textit{School of Computing and Information} \\
% \textit{University of Pittsburgh}\\
% Pittsburgh, United States \\
% yanmuheng@pitt.edu} \\
% \linebreakand
% \IEEEauthorblockN{Yu-Ru Lin}
% \IEEEauthorblockA{\textit{School of Computing and Information} \\
% \textit{University of Pittsburgh}\\
% Pittsburgh, United States \\
% yurulin@pitt.edu}
% \and
% \IEEEauthorblockN{Zian Wang}
% \IEEEauthorblockA{\textit{Department of Computer Science} \\
% \textit{Stony Brook University}\\
% Stony Brook, United States \\
% ziawang@cs.stonybrook.edu}
% }

\IEEEoverridecommandlockouts
\IEEEpubid{\makebox[\columnwidth]{979-8-3503-2445-7/23/\$31.00~\copyright2023 IEEE \hfill} \hspace{\columnsep}\makebox[\columnwidth]{ }}

\maketitle

\IEEEpubidadjcol

\customfootnotetext{$\ddagger$}{These authors contributed equally to this work.}

\customfootnotetext{$\S$}{Corresponding author.}

\begin{abstract}

The escalating food insecurity in Africa, caused by factors such as war, climate change, and poverty, demonstrates the critical need for advanced early warning systems. Traditional methodologies, relying on expert-curated data encompassing climate, geography, and social disturbances, often fall short due to data limitations, hindering comprehensive analysis and potential discovery of new predictive factors. To address this, this paper introduces ``HungerGist'', a multi-task deep learning model utilizing news texts and NLP techniques. Using a corpus of over 53,000 news articles from nine African countries over four years, we demonstrate that our model, trained solely on news data, outperforms the baseline method trained on both traditional risk factors and human-curated keywords. In addition, our method has the ability to detect critical texts that contain interpretable signals known as ``gists.'' Moreover, our examination of these gists indicates that this approach has the potential to reveal latent factors that would otherwise remain concealed in unstructured texts.

\end{abstract}



\section{Introduction}\label{sec:intro}

Africa is at an increased risk of food insecurity due to international and regional conflicts, droughts, climate change, and extreme poverty. In 2022, one in five Africans was at risk of severe hunger, affecting over 140 million people \cite{ReliefWeb_2022}. There is an urgent need for early warning systems in the region that can predict and assess the risk of food crises. 
Such systems will be useful to improve socioeconomic and political stability \cite{Extremeclimateeventsincreaserisk, endtoendassessmentextremeweatherimpacts}, as well as to identify vulnerable areas and effectively target resources to areas of greatest need \cite{buhaug2021vicious, martin2019food}.

Several studies have shed light on the various statistical factors that can be used to predict food insecurity, such as food supply and availability, climates, and agricultural factors \cite{MeasuringHouseholdFoodInsecurityWhy, endtoendassessmentextremeweatherimpacts}. Food insecurity has been studied at a granular level, such as the availability and consumption of food in households \cite{MeasuringHouseholdFoodInsecurityWhy}. Later, researchers began examining macro-level factors such as climate changes or agricultural aspects such as droughts, precipitation, crop growth, vegetation index, land usage, and terrain ruggedness \cite{datadrivenapproachimprovesfoodinsecurity, forecastabilityfoodinsecurity, krishnamurthy2022anticipating}.
% more factors, one by one
Socio-economic elements, on the other hand, are viewed as important as other non-social elements (e.g., agricultural factors) -- in particular, social factors such as international aid, population density, government corruption, economy, and urbanization have been attributed to having a significant impact on the food crisis \cite{FoodsecurityconflictEmpiricalchallenges, Violentconflictexacerbateddroughtrelatedfood, WarConflictFoodInsecurity, ArmedConflictsFoodInsecurityEvidence}.

Understanding the complex interactions between the various factors affecting food insecurity in African countries is crucial for predicting future food crises and preparing resources for crisis mitigation. Existing work, however, heavily relies on the availability of data from a variety of expert-curated sources, including climate, geography, and social conflicts, which are traditionally believed to impact food crises. Yet, these data sources, which are primarily gathered by non-governmental organizations, frequently lack real-time updates, limiting our capacity to predict food crises in a timely manner. Existing works that rely on these resources are unable to adequately account for all possible influences due to the limitations of the available data. This also prevents us from gaining new knowledge and investigating which factors, if any, might help in predicting food crises.

This work seeks to bridge these gaps by leveraging news texts and NLP techniques to create potential solutions for enhancing food crisis prediction. Among the efforts of recent studies to leverage various types of big data, a recent study focused on leveraging news articles, typically published on a daily basis and containing up-to-date information. Using a massive news corpus, they conducted ad-hoc text analysis employing techniques such as topic modeling and keyword statistics \cite{PredictingFoodCrises}. While such analysis sheds light on previously unknown predictive associations between texts and food crises, which are not obtainable via traditional risk factors, the approach still heavily relied on labor-intensive efforts and domain experts' insights.

In this paper, we present \name, a multi-task deep learning model that can both predict and interpret data associated with food insecurity. We observe, based on previous research on food crises, that food crises are frequently influenced by multiple socioeconomic factors, such as social unrest and food price. Thus, we incorporate these into a multi-task model that predicts simultaneously multiple objectives: food price, social instability, and food insecurity. Furthermore, our approach is interpretable and capable of extracting an informative ``gist'' that is predictive of a ``hunger'' crisis. The automatically identified gists represent the most significant contexts and events from a massive corpus of texts, with textual segments indicating a high or low level of food insecurity. Our analysis is based on a corpus of 53,266 news articles collected through the GDELT Project\footnote{GDELT is a database that covers news media in over 100 languages across print, broadcast, and web formats, from almost every corner of each country, on a constant, everyday basis.} \cite{TheGDELT45:online}, covering nine African countries including Uganda, Guinea, Malawi, Mali, Niger, Nigeria, Senegal, Burkina Faso, and the Democratic Republic of the Congo over four years (between 2017 and 2020). Moreover, we propose a data augmenting strategy with a bootstrapping method to increase sample variability in order to deal with data scarcity -- a problem frequently observed in spatial-temporal heterogeneous datasets even when data from certain sources are abundant.

We demonstrate that our model, trained solely on news data, outperforms the baseline method trained on both traditional risk factors and human-curated keywords \cite{PredictingFoodCrises}. This suggests that news corpus alone can be used to improve the ability to predict food insecurity. Moreover, our analysis of model-generated gists reveals that textual segments extracted from news articles that are highly predictive of high or low food crises can provide interpretable signals, demonstrating our model's ability to identify latent factors hidden in unstructured texts. Our key contributions include: 
\begin{enumerate}
\item We present \name, an interpretable predictive model designed to achieve both the prediction of food insecurity and enhanced interpretability. Our model, trained solely on news data, outperforms the baseline approach and is uniquely able to provide textual signals at the sentence level that are highly predictive of food insecurity.
\item We propose a bootstrap data augmentation strategy to deal with the problem of data scarcity frequently seen in spatiotemporal prediction.
\item We design a multi-task learning architecture that incorporates multiple objectives in a prediction framework.  This approach leverages the {\bf complex associations among food insecurity and other potent risk factors including food price and social instability} to improve the prediction of food crises. 
\end{enumerate}

Through extensive experiments, our research demonstrates the potential for data-driven methods to assist government and private sector decision-makers in comprehending complex socioeconomic problems and informing policies to address them effectively.

\section{Related Works}

\subsection{Data-driven approaches in food crisis prediction}
Several decades of research have illuminated a multifaceted realm of diverse factors that exert influence on the issue of food insecurity. In earlier studies, food insecurity was primarily ascribed to the food availability, access, and utilization  \cite{declaration1996rome, MeasuringHouseholdFoodInsecurityWhy, PrioritizingClimateChangeAdaptationNeeds, datadrivenapproachimprovesfoodinsecurity} at the household level -- that is, how households can readily obtain food from the marketplace and have proper ways to consume nutrition. To assess food availability, diverse data sources such as remote sensing data were employed to identify the quantity of local crop production \cite{PrioritizingClimateChangeAdaptationNeeds}.  Contrary to these local factors, the concept of sustainability pertaining to global factors such as environment, economy, and society, has recently been examined as the macro-level factors impacting food security nowadays Among these, the escalation of extreme weather events due to climate change has emerged as a substantial threat to substantial threat to stable crop production \cite{Extremeclimateeventsincreaserisk, endtoendassessmentextremeweatherimpacts}. Political instability and armed conflicts serve as intermediate factors that worsen or are incurred by food insecurity: the limited land use and access due to climate change induce conflict or political unrest \cite{landscapeconflictIDPsaidlanduse, bruck2019reprint}; social conflicts in turn limit people’s mobility and access to marketplaces \cite{Violentconflictexacerbateddroughtrelatedfood, maxwell2021analysing}. This puts restrictions on producers to sell products or inflate food prices \cite{EstimatingFoodPriceInflationPartial}; Consumers have to buy poor-quality foods or skip their meals \cite{WarConflictFoodInsecurity}. Conflicts also force the displacement of large populations, inducing land abandonment and increasing land use around refugees \cite{holleman2017sowing}. These societal-level disruptions can introduce additional shocks to already-vulnerable households in terms of their food consumption \cite{MeasuringHouseholdFoodInsecurityWhy}. Nationwide crisis such as COVID-19 has provided of the potential to threaten millions of people in their food consumption \cite{ImpactsCOVID19pandemicenvironmentsociety, GrainexportrestrictionsCOVID19risk, loopstra2020vulnerability}.

Despite the intricate nature of food insecurity, as evidenced in prior work, where its factors are multidimensional, latent, and intertwined with each other \cite{MeasuringHouseholdFoodInsecurityWhy, datadrivenapproachimprovesfoodinsecurity}, data often heavily relies on data collection efforts undertaken by European or US-based NGOs and governmental organizations. Unfortunately, such sources may not readily provide up-to-date information; Factors that are latent but unavailable from such sources might pose challenges for integration into analysis, despite having the potential of a key driver of food insecurity. Studies are inevitably restricted to only to a few factors, which can harm the predictive power of an underlying model \cite{Machinelearningcanguidefood}.

\subsection{Analysis of large-scale data for predicting food insecurity}
Recent studies have increasingly put efforts into data-driven analysis utilizing large-scale data from a variety of data sources \cite{Predictingpovertywealthmobilephone, Combiningsatelliteimagerymachinelearning, forecastabilityfoodinsecurity} to solve problems on a global scale such as poverty, food insecurity, and climate changes. While only a few studies have utilized large-scale data in food insecurity analysis, the use of textual data noticeably sheds light on the analysis of food insecurity with its prevalence to cope with data scarcity. For example, \cite{FoodinsufficiencyTwitteremotionspandemica} showed that real-time analysis of food insufficiency from Twitter can capture sentiments and emotions correlated with food insufficiency in the US. Other textual data such as YouTube transcription has also been proven to offer qualitative insights into the spatial and temporal characteristics of food insecurity with keywords extracted from topic modeling \cite{Explainingfoodsecuritywarningsignals}. 

Among them, using news stream data has been proven to improve forecasting accuracy and gain a deeper interpretation of the dynamics of food insecurity. Balashankar et al. \cite{PredictingFoodCrises} showed that incorporating a corpus of news articles into the analysis using a set of curated keywords not only improves predictive performance but also complements qualitative evidence that might otherwise be overlooked by traditional features. 
The use of advanced data analytics and machine learning techniques further enhances the ability to detect patterns and predict future trends, making it a powerful tool in the fight against global food crises \cite{barrett2010measuring,shan2014food,bellemare2015rising}.
% \yrl{I have no idea what the blue text tried to do. It only repeats points that have been said and doesn't add anything beyond the original text. Plz remove the blue text and clarify points you tried to argued.}
However, such analysis heavily involves a significant amount of manual effort and requires input from experts, forming a complex pipeline for processing text into useful features. In contrast to prior work, we direct our attention toward a {\it data-driven} approach of capturing a spectrum of events, social issues, and statuses evidenced by texts. By integrating interpretable modules into a deep learning model, our model is capable of extracting key news phrases and events signaled as predictive of a low or high likelihood of food insecurity at both sentence and document levels. This efficacy is demonstrated through qualitative and quantitative evaluations, with scenarios and topics helping better understand factors influencing food crisis.

% \subsection{Language as }

% \subsection{Machine Learning Algorithms}
\section{Problem Statement}\label{sec:problem}

Our goal is to predict the level of food insecurity in each area of interest and automatically identify interpretable signals in unstructured text data to inform the context or events surrounding food insecurity -- that is, the textual {\it gist} of the food insecurity. In this study, we focused on the nine African countries and their food insecurity dynamics on a monthly basis; however, our methodology is applicable for analyzing areas at a different spatial (e.g., districts, regions) or temporal scale (e.g., quarterly or annually), provided that the data are available. 

Suppose there are a set of countries of interest $C$ ($|C|=9$ in our study), and each country $c \in C$ is represented as a collection of its static and dynamic features. The static features, such as population and cropland use, remain the same or change slowly over a longer period of time, and the dynamic features, such as information about rainfall, vegetation, and news reporting,  are updated for each time interval $t$. 

Let $X^{stat}_c$ be the static features of country $c$, and $X^{dyn}_{t, c}$ the set of dynamic features for country $c$ at time $t$. At a future time $t^*$, the level of food insecurity is denoted as a continuous variable $y_{t^*, c} \in \mathbb{R}$ for the country $c$. %\yrl{@Muheng: in your setting, is $y$ discrete ($\mathbb{N}$) or continuous ($\mathbb{R}$)? i.e., are you doing classification or regression?}\mh{it's continuous.}
The outcome variable is derived from the Integrated Food Security Phase Classification (IPC) phase indices. These indices can be accessed through the Famine Early Warning Systems Network (FEWS NET) \cite{HOMEFEWS70:online}, which is further detailed in Sec.~\ref{sec:data-label}.
We transformed the IPC phase indices that are available on a quarterly basis into numerical scores ranging from 1 to 5, with 1 indicating the minimum level of food insecurity (IPC phase: Minimal/None) and 5 the maximal (IPC phase: Catastrophe/Famin). Since the IPC indices are only available on a quarterly basis, we employ linear interpolation to generate a sequence of smoothed food insecurity indices corresponding to each month. This outcome variable $y$ will henceforth be referred to as \fews.

The collection of dynamic features for a country $c$ within an observing \textit{time window} with size $w$ up to time $t$ can be represented as $X^{dyn}_{t-w+1:t,c} = \{X^{dyn}_{t-w+1,c},\ldots,X^{dyn}_{t,c}\}$. Each dynamic feature set may consist of a set of traditional risk factors $X^{dyn,trad}$, which are typically available in tabular form, and a set of textual data $X^{dyn,text}$ organized in a structured or unstructured format, denoted as $X^{dyn}_{t',c}=\{X^{dyn,trad}_{t',c},X^{dyn,text}_{t',c}\}$.

The objective is to forecast \fews $y_{t^*,c}$ for a country $c$ at a future time $t^*=t+\tau$, where $\tau$ is called the {\it lead time}. Given the static and dynamic features of each target country, the forecasting problem can be formulated as learning a function $f(X^{stat}_c, X^{dyn}_{t-w+1:t,c}) \rightarrow y_{t^*,c}$ that maps the static and dynamic features to the \fews at future time $t^*$ for each {\it target} country $c$. In our study, we are only interested in the short-term dynamics with a focus on $\tau=1$ and $w=1$; however, our methodology is flexible to analyze a larger time window size and a longer lead time (i.e., $\tau\ge1$ and $w\ge1$). To facilitate the detection of predictive and informative signals, we aim to develop a model that is not only capable of accurately predicting the future \fews, but also of differentiating the contribution of the textual features, i.e. the most predictive textual segments.

\section{Data}
\label{sec:data}
In our analysis, we leverage multiple types of data collected from different sources. This includes the level of food insecurity as an outcome indicator of food insecurity level (Section \ref{sec:data-label}), and traditional risk factors such as climate, agricultural, and conflict-related factors (Section \ref{sec:data-traditional}) and textual data (Section \ref{sec:data-textual}) that potentially capture hidden signals that are predictive or interpretable for food instability. In the analysis, we conduct food crisis prediction at the country-level analysis due to news data collected at the country-level as its geographical granularity (Section \label{sec:data-textual}), In the analysis, we focus on nine African countries with all data between 2017 and 2020 available from data collection described in the following sections. 

\subsection{Food security phase}
\label{sec:data-label}
We employ IPC phase available from the FEWS NET, a leading provider of early warning and analysis on acute food insecurity around the world, in our analysis as our target variable. The IPC phase provides a standardized scale for classifying the severity of food insecurity over households in areas as five distinct phases, including (1) Minimal/None, (2) Stressed, (3) Crisis, (4) Emergency, and (5) Catastrophe/Famine. The data encodes the food insecurity across 37 countries over the period on a quarterly basis at the global, regional and country levels. As described in Sec.~\ref{sec:problem}, the quarterly ordinal indices obtained from the IPC phase data are transformed into a country-level, numerical \fews at a monthly basis.

\begin{table}[]
\small
\centering
\caption{\label{tab:data_traditional} \textbf{List of traditional risk factors.}}
\begin{tabular}{@{}ll@{}}
\toprule
\textbf{Type}  & \textbf{Feature}                       \\ \midrule
Time-varying   & Rainfall index                         \\
               & Normalized difference vegetation index \\
               & Food price index                       \\
               & Conflicts and fatalities               \\
               & Terrain ruggedness index               \\ \midrule
Time-invariant & District size                          \\
               & Share of cropland use                  \\
               & Share of pasture use                   \\
               & Population count                       \\ \bottomrule
\end{tabular}
\end{table}

\subsection{Traditional risk factors}
\label{sec:data-traditional}

We have identified nine traditional risk factors that have been widely hypothesized to influence food insecurity in previous studies \cite{Finegrainedpredictionfoodinsecurityusing, datadrivenapproachimprovesfoodinsecurity, PredictingFoodCrises}, as listed in (Table \ref{tab:data_traditional}). These individual features present multidimensional aspects of country-level information in a time-invariant or time-varying manner on a monthly basis, which are related to geography, climates, and social stability such as conflict events as listed in Table \ref{tab:data_traditional}.

\subsection{Texts from news data}
\label{sec:data-textual}
Our textual features were extracted from texts in news articles collected from GDELT Project. The tool allows us to retrieve articles under certain themes, e.g., “food security” and “agriculture” in our data collection. Based on the data availability from the tool, our news corpus covers news articles from 9 countries over the period from January 2017 to December 2021. For each country, we only collected news data written in English or in the country’s official language such as French. 
After gathering the raw news data, we translated all non-English articles to English using the Google Translate service. This step was necessary because our preliminary tests revealed that the pretrained language model for French did not perform as well as the English language model, nor as effectively as the translated texts. The English language model's superior performance drove our decision for this approach, which we will discuss further in the Discussion section.
% After collecting the raw news data, we used Google translate service to convert all of the non-English news articles into English\footnote{we tested the pretrained language model in French but it under-performs English language model and translated texts}. 

% \yrl{@Muheng to add a footnote to justify the use of translated text}

% \yrl{@Muheng: is the following paragraph correct? Shall we move LDA to the result section and the DistilBert to method section?}

% From this corpus, we extracted topical embeddings of articles driven by two topic models as textual features to obtain meaningful article-level representations as model input: 1) LDA-based article embedding: The article texts were aggregated for each month, and preprocessed through lemmatization and n-gram tokenization. These tokens were modeled by Mallet LDA, with the elbow rule to determine the number of topics. 2) distill-BERT-based article embedding: The news articles were also aggregated per month and fed into distill BERT. Each article is represented with a combination of two vectors from the models. This feature only includes countries of Burkina Faso, Guinea, Mali, Nigeria, and Senegal.

\section{Methods}

Our objective is to forecast \fews for each country at a given time $t+1$ using data observed at $t$. There are two technological obstacles: (1) Limitations on samples: In contrast to the large set of input data, which contains numerous features and abundant texts, the output indices are relatively sparse: for each month we have only one single food crisis score for each of the nine countries, therefore there are only nine data points at each time step. The limited outcome indices severely restrict the number of samples that can be used to develop a predictive model. (2) Complex risk associations: It has been known that food insecurity is intricately linked to other social and economic risks \cite{loopstra2020vulnerability, higashi2017family}. Thus, food insecurity co-occurring with complex socioeconomic risks may not be fully captured by a single variable, \fews. We propose solutions for overcoming these two technical challenges. First, we propose bootstrap techniques to increase samples with data variability (Sec.~\ref{sec:bootstrap}). Second, we propose multi-task modeling to incorporate signals learned from the most pertinent social and economic risk factors (Sec.~\ref{sec_034:multi-task-model}).

\begin{figure*}[h]%
\centering
%\small
\includegraphics[width=\textwidth]{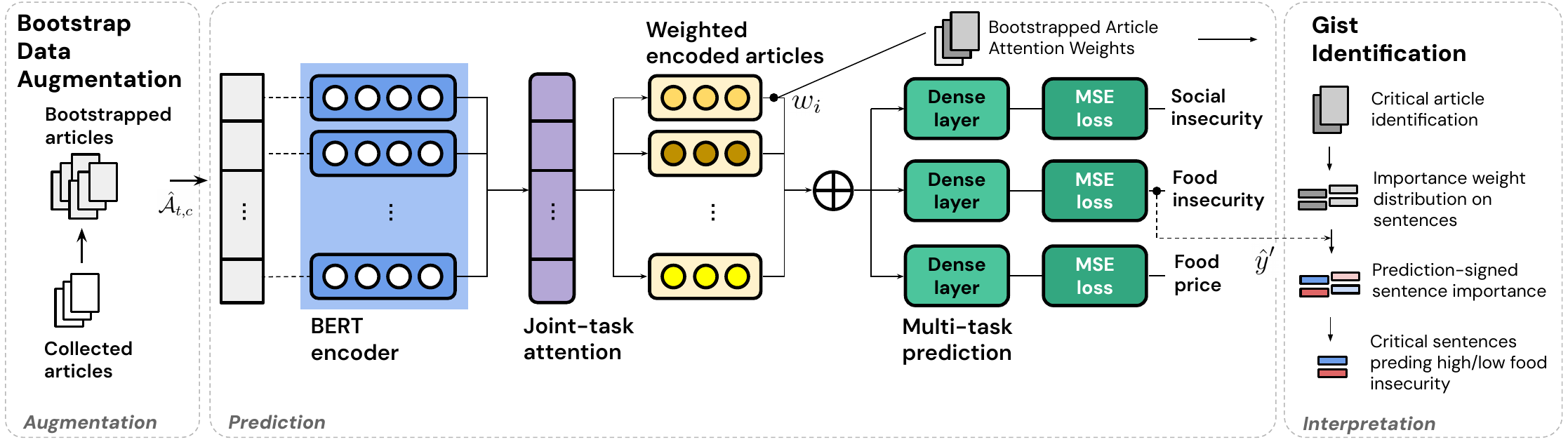}
\vspace{-1em}
\caption{{\bf The deep neural architecture of \name} is an interpretable multi-task classifier for predicting food insecurity (\fews) and two other related risk factors, social insecurity and food price. Our model consists of three parts to tackle the essential challenges in food crisis predictions: 1) Augmentation: Our bootstrap data augmentation strategy tackles data scarcity in spatiotemporal datasets; 2) Prediction: Recognizing food insecurity as a complex association of multiple objectives, our model is configured for multi-objective prediction to leverage multiple signals including social insecurity, food insecurity and price; 3) Interpretation: Sentences as highly indicative of high or low food insecurity, extracted as gists from an interpretable module in the deep neural model, are readily available to support the sense-making of high or low food crisis.}
\label{fig:teaser}
\end{figure*}

\begin{table}[]
\centering
\small
\caption{\label{tab:news_data} \textbf{Overview of news data.}}
\begin{tabular}{@{}lll@{}}
\toprule
\textbf{Country} & \textbf{\# Articles} & \textbf{Language} \\ \midrule
All              & 53,266               & English, French   \\
Uganda           & 7,086                & English           \\
Congo            & 3,500                & English           \\
Guinea           & 3,983                & English, French   \\
Malawi           & 5,945                & English           \\
Mali             & 18,034               & English, French   \\
Niger            & 2,112                & English, French   \\
Nigeria          & 8,030                & English           \\
Senegal          & 3,500                & English, French   \\
Burkina Faso     & 4,576                & English, French   \\ \bottomrule
\end{tabular}
\end{table}

\subsection{Bootstrap Data Augmentation}\label{sec:bootstrap}
For each country $c$ at a given time $t$, a collection of news articles, $\mathcal{A}_{t,c}$, is gathered and can be used as input for the prediction. 
Since each article set corresponds to only a single outcome variable ($y_{t+1,c}$), this would result in limited samples and low data variability. We therefore propose the following bootstrap sampling strategy to increase both data sample size and variability, as well as to facilitate the identification of interpretable and informative text segments.

Let $\mathcal{A}_{t,c}=\{A_1, A_2, ...\}$ be the collection of news articles for the country $c$ published at the time $t$, where $A_i = \{s_1, s_2, ..., s_{n_i}\}$ represent an article $i$ consisting of $n_i$ sentences. For each observed article collection $\mathcal{A}_{t,c}$, we create a set of {\it pseudo-collection}, $\{\hat{\mathcal{A}}^{(1)}, \hat{\mathcal{A}}^{(2)}, …, \hat{\mathcal{A}}^{(K)}\}_{t,c}$ as $K$ different textual inputs that correspond to the same set of other types of features and the same outcome as $\mathcal{A}_{t,c}$'s.

Each pseudo-collection $\hat{\mathcal{A}}$ is created to have a fixed number of $m$ {\it pseudo-articles}, i.e., $\hat{\mathcal{A}}=\{\hat{A}_1, \hat{A}_2, ..., \hat{A}_m\}$, where each pseudo-article $\hat{A}_i$ is created by drawing a fixed number of $n$ sentences from the corresponding observed collection $\mathcal{A}_{t,c}$. We consider bootstrapping at the sentence level because this level of signals will aid in the interpretation of the food insecurity context -- as opposed to articles, which may contain multiple subtopics and be of varying lengths. In our study, $n$ and $m$ were empirically determined to be $21$ and $85$, respectively, based on the median number of sentences per article and the median number of articles per country, per month observed in our dataset. In our experiment, various pseudo-sets are used as training inputs. Therefore, in a 10-fold cross-validation setting, the total number of samples for the nine countries is 9 (countries) $\times$ 44 (months) $\times$ 10 (folds) $= 3960$ samples, resulting in 336600 (3960 $\times$ 85) pseudo-articles.

\subsection{Multi-task Modeling} \label{sec_034:multi-task-model}
In order to address the complex socioeconomic risks that are not fully captured by the single supervised signals obtained from \fews, we propose multi-task learning that can simultaneously predict \fews and potent socioeconomic risks. Such a side-task or multi-task modeling approach has been demonstrated to be effective in other domains \cite{ahn2022tribe}. Within the context of food security, two factors, food price and social insecurity status, have been identified as having the strongest relationship with food insecurity \cite{FoodsecurityconflictEmpiricalchallenges, Violentconflictexacerbateddroughtrelatedfood}. Therefore, we propose the following architecture for multitask learning that simultaneously predicts three distinct types of outcomes, \fews, food price, and social insecurity level, denoted as $Y_{t^*,c}=\{y^{crisis},y^{price},y^{social}\}_{t^*,c}$ for time $t^*$ and country $c$. %\yrl{@Muheng: are the three outcome variables in the same temporal scale? or do you use fine-grained, e.g. daily values, for food prices and food insecurity?} \mh{they are all at monthly level}

We develop an end-to-end multi-task attention neural network to predict the joint outcome with a set of traditional risk factors (dynamic and static) and text data as input. The attention mechanism \cite{AttentionInterpretable} is employed to identify the predictive and informative textual signals in order to facilitate interpretation. Note that the textual data has been generated by the bootstrap sampling described in Sec.~\ref{sec:bootstrap}. For each given pseudo-article $\hat{A}_i$, we use a pre-trained DistilBERT model \cite{sanh2019distilbert} to generate a text embedding. Let $\vec{h_i} \in \mathcal{R}^d$ be the text embedded vector for $\hat{A}_i$. In the following, we omit the sub-indices $t$ and $c$ and simply use $Y_j$ as the predicting output for task $j$, and we represent the text input as $\vec{h}_A$ as a weighted sum of $\{\vec{h_i}\}$. The article-specific weight $w_i$ is learned through the shared layers and attention mechanism. 

The shared layers with parameter $\theta$ are given by $g_{\theta}(\{\vec{h_i}\})$, and the attention mechanism produces the (scaled) attention weights $\{w_i\}$ for each pseudo-article. Then, the attention weights are used to create the aggregated textual representation for each country and time:
\begin{equation}
    \vec{h}_A = \sum_{1}^{m} w_i \vec{h_i}.
\end{equation}

Next, we use the aggregated textual representation as the input to obtain the task-specific function for task $j$ with parameters $\theta_j$ that produce output:
\begin{equation}
   \hat{Y}_j = f_{\theta_j}( \vec{h}_A).
   % , \{X^{stat}, X^{dyn,trad}\} ).
\end{equation}
%\yrl{@Muheng: I'm not sure how you use the traditional features so I make this up. Can you clarify and correct this part if its' incorrect?}\mh{The deep model does not use traditional features.}

For each task $j$, the loss function is given by $\mathcal{L}_j(Y_j, \hat{Y}_j)$, and we consider the MSE loss: 
    $\mathcal{L}_j = \sum({\hat{Y}_j - Y_j})^2$.
Thus, the overall loss for multi-task learning is the sum of the individual task losses:
\begin{equation}
    \mathcal{L} = \mathcal{L}_{crisis} +  \mathcal{L}_{price} +  \mathcal{L}_{social}.
\end{equation}
Fig.~\ref{fig:teaser} illustrates the proposed deep learning architecture.

\subsection{Interpretable Analysis}\label{sec:interpret}
We use the attention layer output weights $w_i$ (corresponding to a pseudo-article $\hat{A}_i$) to analyze the textual segments that are highly predictive of high and low \fews. These informative textual segments, which we term {\it gist}, are obtained as follows. 

First, the importance of each pseudo-article is determined by its \fews prediction score, which is calculated as $\dot{w}_i = w_i\hat{y}'$,, where $\hat{y}'$ is the min-max normalization of the prediction outcome $\hat{y}$. The attention weights $w_i$ are mapped to a zero-centered scale in which a greater value indicates a stronger association with positive predictions and vice versa.
Second, to obtain sentences that are important for the prediction, we distribute the weights uniformly across bootstrapped $n$ sentences within each article, as $w_{s} = \frac{\dot{w}_i}{n}, \forall s \in \hat{A}_i$. The importance score $w_s$ for sentence $s$ is then used to rank all sentences and identify those with high or low \fews.

\section{Experiment Setup} \label{sec_035:experiment-setup}
\begin{table*}[]
\centering
\small
\caption{\label{tab:regression_results} \textbf{Prediction Results. The RMSE score of predictions from models in comparison.}}
\begin{tabular}{@{}lccccc@{}}
\toprule
 &
  \multicolumn{1}{l}{\textbf{Baseline}} &
  \multicolumn{1}{l}{ \parbox{2.4 cm}{\centering \textbf{Single Task (Food Crisis)} } } &
  \multicolumn{1}{l}{ \parbox{2.4 cm}{\centering \textbf{Double-task with Food Price} } } &
  \multicolumn{1}{l}{ \parbox{2.4 cm}{\centering \textbf{Double-task with Social Insecurity} } } &
  \multicolumn{1}{l}{\textbf{Triple-task}} \\ \midrule
All          & 1.074          & 0.819 & 0.841 & 0.827          & \textbf{0.790} \\ \midrule
Uganda       & 0.922          & 0.914 & 0.862 & 0.961          & \textbf{0.798} \\
Congo        & 0.982          & 1.036 & 1.040 & 1.036          & \textbf{0.949} \\
Guinea       & 1.660          & 0.418 & 0.444 & \textbf{0.412} & 0.548          \\
Malawi       & 1.055          & 0.667 & 0.704 & 0.769          & \textbf{0.662} \\
Mali         & 1.278          & 1.041 & 1.096 & 0.995          & \textbf{0.982} \\
Niger        & \textbf{0.636} & 1.066 & 1.125 & 0.999          & 1.058          \\
Nigeria      & 1.122          & 0.538 & 0.459 & 0.564          & \textbf{0.403} \\
Senegal      & 1.266          & 0.331 & 0.370 & \textbf{0.337} & 0.452          \\
Burkina Faso & \textbf{0.743} & 0.942 & 1.018 & 0.987          & 0.94           \\ \bottomrule
\end{tabular}
\end{table*}

{\bf Prediction Task.}
For the multi-task model, we use the above-reported deep network to take the batches of articles for each country-month tuple and give the predictions of the next month.
For each article, each sentence is embedded with a pre-trained BERT model \cite{sanh2019distilbert}, producing a sentence vector $\vec{s} \in \mathcal{R}^{768}$ (the dimension is decided by the BERT model architecture). 
% \yrl{@Muheng: can you clarify which BERT is used? is it DistilBert as in the method section?}
Then each article is represented by the average vector of the sentences it composed, producing an article embedding $\vec{h} = \overline{\vec{s}}$ of the same dimension.
These article vectors are input to the network together per country-month pair.

We create a hold-out experiment setting aligned with earlier studies \cite{PredictingFoodCrises}.
The country-month having dates after (including) January 2020 are in the test set.
The other tuples with dates after April 2019 are collected as development sets for model tuning and parameter selection.
All other tuples from January 2017 to March 2019 are used as training set.
This split results in 2340 samples in the training set, 720 samples in the development set, and 900 samples in the testing set.

We use equal weight over the three model tasks.
The model is evaluated on the RMSE on \fews every five training steps.
If the RMSE does not optimize for more than ten evaluation steps, the training session stops.

% \paragraph{Interpretable Analysis.}
% We use the attention layer outputs $w_i$ (corresponding to article $\hat{A}_i$ in a country-month tuple) to explore textual features that associate with high and low \fews.
% After the model training, each pseudo-article $\hat{A}_i$ is assigned an attention weight $w_i$.
% We first multiply the the weight $w_i$ by the min-max centralized \fews prediction score as $\dot{w}_i = w_i\hat{y}$, where the attention weight are mapped to a zero-centered scale where larger value indicates higher association with positive \fews predictions and vice versa.
% We then evenly distribute the weight among bootstrapped $n$ sentences within each article $w_{s} = \frac{\dot{w}_i}{n}, \forall s \in \hat{A}_i$.
% We rank the sentences on the calculated $w_s$ to identify sentences associated with high or low \fews.

{\bf Baseline Model.}
To evaluate the predictive performance, we compare our model, \name, with a baseline model described in \cite{Finegrainedpredictionfoodinsecurityusing}\footnote{
Our study replicates the model originally published in \cite{Finegrainedpredictionfoodinsecurityusing} to serve as a baseline for comparison. Subsequently, the author of \cite{Finegrainedpredictionfoodinsecurityusing} released an updated version, which employs a random forest algorithm trained on quarterly data. In our work, we establish a new baseline model designed to predict outcomes on a monthly basis, and conduct experiments using a random forest model and present those results.}
% Note that the model published in \cite{Finegrainedpredictionfoodinsecurityusing} was replicated as the baseline model for our paper during the time of our study. Later, the same researcher published an updated version of the model, with a random forest model trained on quarterly observations. In this work, we create a baseline model for prediction on a monthly basis for comparison, and we report the experiment results with random forest model.}. 
In this study, Balashankar et al. \cite{PredictingFoodCrises} proposed a simple machine-learning framework that utilizes both conventional risk features and news articles as input. All features are incorporated into a panel autoregressive distributed lag (ADL) model that includes both time-lagged temporal factors and time-invariant factors as inputs for prediction.

In particular, the baseline model contains three types of features: (1) self-lagged target features for three to eight months; (2) time-varying features with three to eight months of lagged values; and (3) time-invariant features. The time-varying features consist of four traditional factors listed in Table \ref{tab:data_traditional} and keyword-based textual features derived from news data. Following their work \cite{Finegrainedpredictionfoodinsecurityusing}, the keyword-based features are created by first curating a fixed list of keywords provided by domain experts, and then obtaining the frequencies of these pre-defined keywords to generate a tabular form of textual features. 
%We train the baseline models using cross-validation on split data. We partitioned the observational months into 10 periods as folds that progressively span recent months. Within each fold, we subsequently segmented the data into training, validation, and testing subsets using a temporal split.

\section{Results}

\subsection{Food Crisis Prediction}\label{sec:prediction}

In Table \ref{tab:regression_results}, we report the experiment results for prediction performance as measured by the Root Mean Square Error (RMSE). In addition, we evaluate the performance of simplified versions of our complete multi-task model, including a single-task setting (predicting \fews only) and two dual-task settings (also predicting food price or social insecurity). The triple-task setting (considering all three losses) uses our complete multi-task model. In general, the results indicate that all four configurations of our modeling architecture produce better prediction results than the baseline model, which relies on a combination of traditional and keyword-based features. Specifically, the complete triple-task model reduces RMSE by 26\% (from 1.074 to 0.790) and outperforms single- and dual-task models for the majority of countries and overall. As indicated by the significant performance improvement over the baseline approach, our proposed architecture is able to leverage unstructured text input effectively, without relying on a predefined keyword list or labor-intensive feature engineering effort.

We also observe that there is variation in country-level prediction outcomes. Our triple-task model achieves a lower RMSE in the majority of countries compared to other model variants. In Guinea and Senegal, however, the dual-task model for predicting \fews and social insecurity achieves a lower RMSE than the complete model. In Niger and Burkina Faso, the performance of the baseline model is better than our model variants. This suggests that while our models trained on news data demonstrate competitive performance, in certain countries, traditional risk factors and curated keywords still provide more crucial information than news data for food insecurity prediction.

\subsection{Gist of Food Crisis}\label{sec:interp_result}

\begin{figure*}[ht!]
\centering
\includegraphics[width=0.95\textwidth]{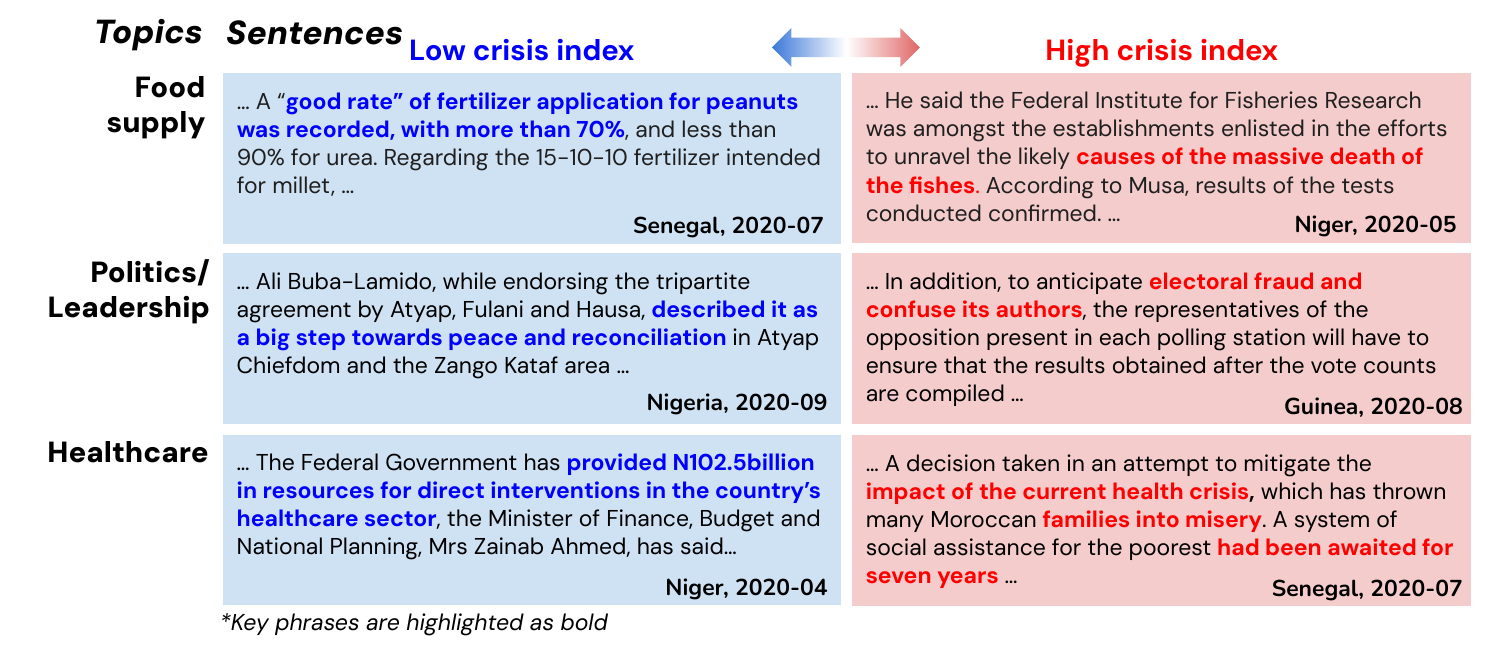}
\vspace{-1em}
\caption{\label{fig:examples} 
{\bf Examples of identified gists} corresponding to the high and low food crisis indices are highlighted as red or blue, respectively, in the articles. These sentence-level evidence allows a nuanced analysis of food insecurity with fine-grained details of events and contexts  related to food crisis across diverse topics.}
\end{figure*}

Using the method outlined in Sec.~\ref{sec_034:multi-task-model}, we determine the {\it gists} -- the informative sentences most strongly associated with high or low \fews. We highlight how these sentences contribute to an understanding of the context of food insecurity. 

\begin{figure}[ht!]
\centering
\includegraphics[width=\columnwidth]{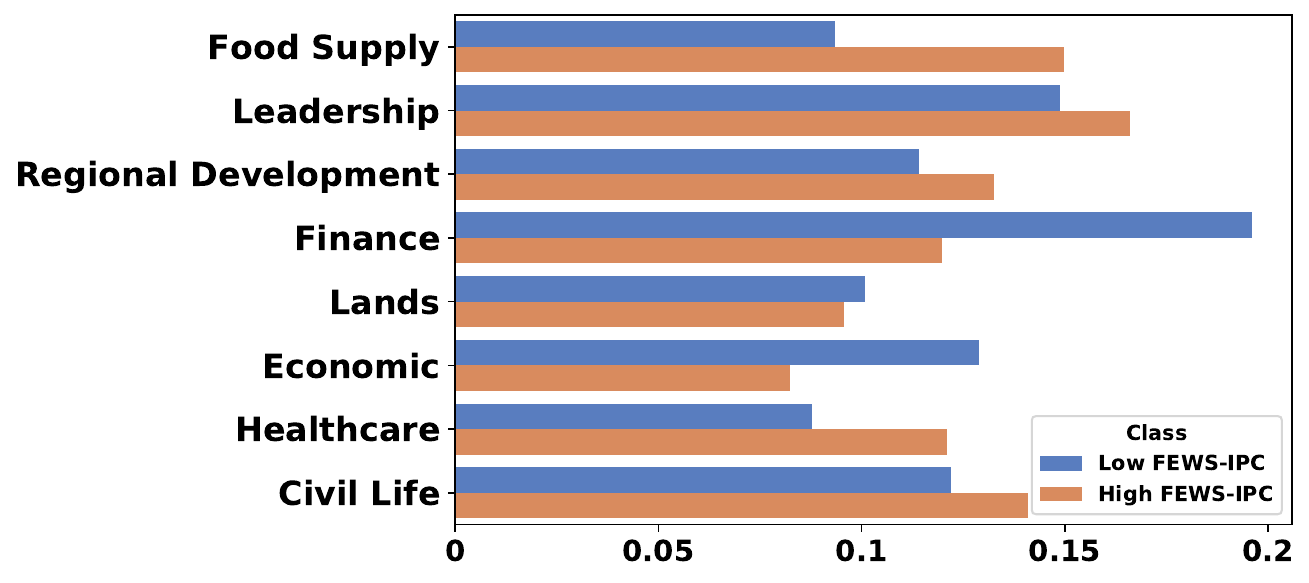}
\vspace{-2em}
\caption{\label{fig:topic} 
\textbf{Topic distribution of identified gists.} The model-generated dual-sided interpretations with sentence-level importance scores allow a nuanced contrast to distinguish factors more closely associated with either a high or low risk of food insecurity. Our topic analysis specifically highlights Finance/Economic Development as a determinant highly indicative of low food insecurity, while Food Price/Health Care is identified as strongly associated with high food insecurity.
% \yrl{is the figure quality improved? is the color scheme print-friendly?}
}
\end{figure}
\subsubsection{Topic Analysis of Important Sentences}
We use a topic modeling approach to analyze the topics of the extracted important sentences. The most important sentences are determined as the top 5\% and bottom 5\% according to the sentence weights $\{w_s\}$. We apply Mallet LDA \cite{graham2012getting} on the entire text corpus. The texts are lemmatized and stopwords are removed before the topic modeling.
We discover a total of eight topics and manually assign names to them according to the top keywords of each topic. The topics are: Food Price, Leadership, Regional Development, Finance, Land Scale, Economic Development, Health Care, and Civil Life.
The trained topic model is used to infer the topic probability for each of the extracted important sentences. 

In Fig.~\ref{fig:topic}, we show the distribution of topics among the most important sentences. On almost all topics, we observe distinguishable differences between sentences that predict low and high \fews. Sentences corresponding to low \fews appear to be more related to Finance and Economic Development, whereas sentences corresponding to high \fews appear to be more related to Food Price and Health Care. This observation is consistent with previous research, which found that food price and other social issues are typically the most important determinants of food insecurity \cite{conceiccao2009anatomy, FoodsecurityconflictEmpiricalchallenges}.

\subsubsection{Semantic Differences of Extracted Gists between High and Low Crisis Index}

In Fig.~\ref{fig:examples}, we present a list of example sentences corresponding to the high and low \fews in three different topics. We extract the most important sentences for each topic based on the sentence weights $w_s$. We find that texts that are indicative of a low crisis index are more likely to make reference to ``good'' events, while texts that are indicative of a high crisis index are more likely to be linked to ``bad'' events.

For the topic ``Food Supply,'' the example reports positive fertilizer application rates for peanut farms, which resulted in a signal in the deep learning model that predicts a low crisis index outcome. Another example within the same topic, on the other hand, reports on an investigation into a massive fish death.  Similarly, for the topic ``Politics/Leadership,'' an announcement by political figures on peaceful reconciliation corresponds to a low crisis index, whereas a report on election polling fraud predicts a high food crisis.  Finally, in ``Healthcare,'' the national budget funding healthcare section is useful for predicting a low crisis index, whereas a health crisis caused by delayed social assistance is useful for predicting a high crisis. 

This provides evidence that our proposed model can effectively identify important context and news events that correlate with food crisis forecasts in Africa.

\section{Discussion and Conclusion}\label{sec:discussion}

This paper introduces \name, an interpretable predictive model for food insecurity analysis. We demonstrated that our multi-task deep learning model, tasked with simultaneously predicting three critical objectives (food insecurity, food price, and social insecurity) not only improves food crisis prediction but also enhances interpretability, as evidenced by our experiment results and analysis. Compared with previous works \cite{forecastabilityfoodinsecurity, PredictingFoodCrises, gholami2022food}, our model uniquely provides interpretability by detecting sentence-level textual signals useful for understanding the context associated with food crises. In the experiment results, we demonstrated that the most significant sentences, as identified by our interpretable model, provided meaningful contextual information associated with food insecurity -- for example, the use of fertilizer to increase crop production and reconciliation among different leaderships correspond to a lower food crisis situation, whereas electoral fraud and health crises correspond to a higher food crisis situation. 

Our proposed method can be used as part of an early warning system's analysis pipeline to (1) reduce the manual feature engineering efforts and intensive expert inputs typically required by prior approaches and (2) uncover novel insights in news data that may be associated with future food insecurity, which is not possible with existing works. We anticipate that analysts and decision-makers in governments, non-profit organizations, and international government bodies will benefit from the proposed data-driven approach, which will enhance their ability to adapt to rapidly changing conditions caused by local and global events such as the COVID-19 pandemic \cite{falkendal2021grain}.

This work has several limitations. While our analysis is conducted at the country level, a more granular level of analysis (e.g., at the district level) may better capture the variation of the food crisis occurring in local areas. With available data (e.g., identifying local news sources), our analysis can be adapted for use at various spatial-temporal scales. 
In this work, we use translated texts from French-language news articles. This is empirically determined as we found that translated text with an English language model performs better than original texts with a French language model of the same model size. This does not imply that the translated text input works better, as this could be due to the higher overall quality of English model through pretraining. Future research could look into this further with a better available French language model.
In this work, we built upon the BERT model family, given its superior performance over conventional NLP methodologies \cite{sanh2019distilbert}, such as keyword-based and other earlier neural network models. The proposed framework, however, is not limited to the BERT family and can be easily extended to include other LLMs \cite{katz2023gpt}.
Further, our analysis can be expanded to incorporate conventional risk factors to improve both predictive performance and interpretability. As discussed in the results of our experiment, features from various sources have distinct advantages to enhance the ability to predict food crises. To further improve the interpretability of textual signals, future research may also incorporate traditional risk factors. Additionally, the results of our study suggest that the baseline methodology demonstrates superior performance in two of the countries being analyzed. 
By extending our analysis to include additional variables, such as traditional risk factors, or incorporating a hierarchical spatiotemporal deep learning approach \cite{ertugrulCASTNetCommunityAttentiveSpatioTemporal2019}, we could gain more in-depth understanding of the unique features of these countries. This expanded analysis might explain why certain benefits were observed exclusively in other countries.
Finally, it should be noted that the current study does not explicitly evaluate the quality of the news articles that were included in the analysis. Given the growing concern about the prevalence of misinformation around the world, especially in times of crisis (such as the COVID-19 pandemic \cite{tengCharacterizingUserSusceptibility2022}), it is imperative for future research to account for news source quality in order to reach more reliable conclusions.

\section*{Acknowledgment}
The authors would like to acknowledge the support from AFOSR awards and DARPA Habitus program. Any opinions, findings, and conclusions or recommendations expressed in this material do not necessarily reflect the views of the funding sources.

\bibliography{ref, references-yongsu}
\bibliographystyle{IEEEtran}

% \vspace{12pt}
% \color{red}
% IEEE conference templates contain guidance text for composing and formatting conference papers. Please ensure that all template text is removed from your conference paper prior to submission to the conference. Failure to remove the template text from your paper may result in your paper not being published.

% \input{99_appendix.tex}

\end{document}